\documentclass[letterpaper, 10 pt, conference]{ieeeconf}  % Comment this line out if you need a4paper

\IEEEoverridecommandlockouts  % This command is only needed if you want to use the \thanks command

\usepackage{graphics} % for pdf, bitmapped graphics files
\usepackage{epsfig} % for postscript graphics files
\usepackage{times} % assumes new font selection scheme installed
\usepackage{amsmath} % assumes amsmath package installed
\usepackage{amssymb}  % assumes amsmath package installed

\usepackage{booktabs}
\usepackage{multirow}
\usepackage{makecell}
\usepackage{amsfonts}
\usepackage{comment}
\usepackage[style=base]{caption}

\usepackage[pagebackref=true,breaklinks=true,colorlinks,bookmarks=false]{hyperref}
\usepackage[capitalize,nameinlink]{cleveref}

\title{\LARGE \bf
A generic diffusion-based approach\\for 3D human pose prediction in the wild
}

\author{Saeed Saadatnejad$^{1}$, Ali Rasekh$^{2}$,
Mohammadreza Mofayezi$^{2}$,
Yasamin Medghalchi$^{2}$,
Sara Rajabzadeh$^{2}$, \\
Taylor Mordan$^{1}$ and
Alexandre Alahi$^{1}$% <-this % stops a space
\thanks{$^{1}$EPFL, Lausanne, Switzerland  (e-mail: saeed.saadatnejad at epfl.ch)}%
\thanks{$^{2}$The research was conducted during an internship at EPFL}%
}

\begin{document}
\maketitle
\thispagestyle{empty}
\pagestyle{empty}

%===============================================================================

\begin{abstract}
Predicting 3D human poses in real-world scenarios, also known as human pose forecasting, is inevitably subject to noisy inputs arising from inaccurate 3D pose estimations and occlusions.
To address these challenges, we propose a diffusion-based approach that can predict given noisy observations. We frame the prediction task as a denoising problem, where both observation and prediction are considered as a single sequence containing missing elements (whether in the observation or prediction horizon). All missing elements are treated as noise and denoised with our conditional diffusion model. To better handle long-term forecasting horizon, we present a temporal cascaded diffusion model. 
We demonstrate the benefits of our approach on four publicly available datasets (Human3.6M, HumanEva-I, AMASS, and 3DPW), outperforming the state-of-the-art. Additionally, we show that our framework is generic enough to improve any 3D pose prediction model as a pre-processing step to repair their inputs and a post-processing step to refine their outputs.
The code is available online: \url{https://github.com/vita-epfl/DePOSit}.
\end{abstract}

%===============================================================================

\section{Introduction}

Robots and humans are poised to work in close proximity. Yet, current technology struggles to read and anticipate the motion dynamics of humans. Predicting 3D human poses enables a safe co-existence between humans and robots, with direct applications in social robotics~\cite{chen2019crowd}, autonomous navigation~\cite{mangalam2020disentangling}, assistive robotics~\cite{wagner2018targeted, wenger2016spatiotemporal}, and human-robot interaction~\cite{butepage2018anticipating, lasota2017multiple}.

Predicting a sequence of future 3D poses of a person given a sequence of past observed ones, also referred to as human pose forecasting, is a challenging task since it must combine spatial and temporal reasoning to output multiple plausible outcomes.
Previous models have yielded satisfactory results~\cite{mao2020history, ma2022progressively}, yet they fail to produce acceptable outcomes in noisy settings.
Minor offsets from detection methods or partial occlusions of body parts can drastically impact the prediction accuracy.

\begin{figure}
    \centering
    \includegraphics[width=0.9\columnwidth]{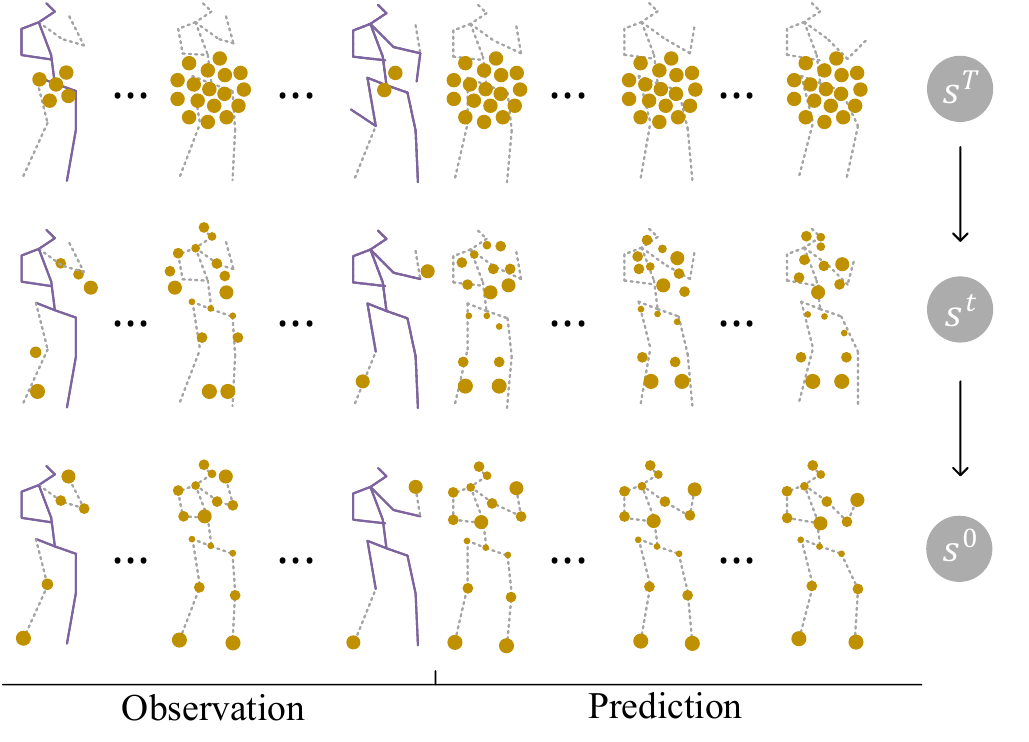}
    \caption{Our proposed conditional diffusion model denoises the input sequence $s^T$ over $T$ steps by simultaneously 1) predicting poses for the future frames and 2) repairing the noisy observations in the case of partial occlusion (first column), missing whole frame (second column), or inaccurate observations (third column). The large yellow circles depict the Gaussian noise we consider for unavailable joints, which gradually become smaller and fit into the correct locations.}
    \label{fig:overview}
\end{figure}

Denoising Diffusion Probabilistic Models (DDPMs)~\cite{ho2020ddpm} are one type of generative models that can denoise input signals iteratively. Motivated by this property, we propose a diffusion model that explicitly handles noisy data input so that it not only predicts accurate and in-distribution poses, but can also be used in the wild.
As depicted in \Cref{fig:overview}, we construct a full sequence of observation and future frames where noise is placed in the missing observation elements and future poses.
Our model denoises this sequence in several steps and produces the correct predictions.
Naively predicting all future frames simultaneously results in inaccurate predictions in later frames. Hence, we propose a model comprised of two temporally-cascaded diffusion blocks. The first block predicts the short-term poses and repairs the noisy observations (if applicable), while the second block uses the output from the former as a condition to predict the long-term poses.
We also leverage our model in a generic framework that can improve the performance of state-of-the-art prediction models in a black-box manner. To this end, we use our diffusion-based model as a pre-processing step to repair the observations providing pseudo-clean data for the prediction model to make more reliable predictions. 
Our model can then be used as a post-processing step to further refine these predictions.

To summarize, our contributions are three-fold:
\begin{itemize}
    \item We frame the 3D human pose prediction task as a denoising problem.
    \item We propose a two-stage diffusion model outperforming the state-of-the-art in both clean and noisy observation settings.
    \item We introduce a generic framework that leverages our model through pre-processing (repairing the input) and post-processing (refinement), which can enhance any pose prediction model.
\end{itemize}

\begin{figure*}
    \centering
    \vspace{5pt}
    \includegraphics[width=0.75\textwidth]{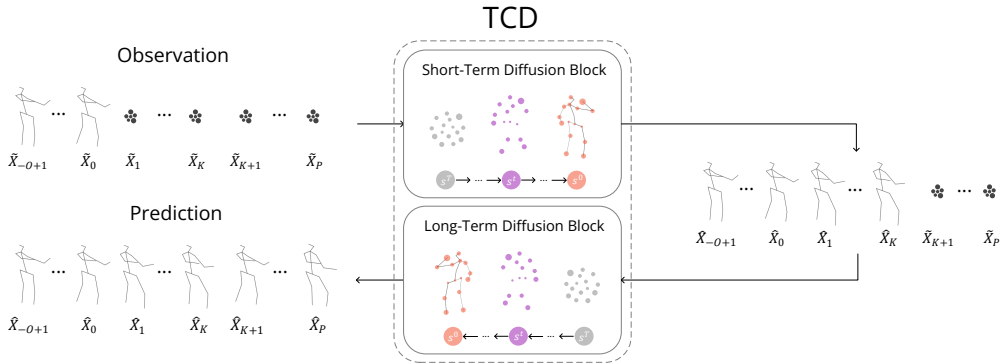}
    \caption{Overview of our Temporal Cascaded Diffusion (TCD). The short-term diffusion block (top) takes the observed sequence padded with random noise and predicts short-term human poses in $K$ frames.
    The predicted sequence along with the observation padded with random noise is given to the long-term diffusion block (bottom) to predict for all $P$ frames.}
    \label{fig:model}
    \vspace{-2pt}
\end{figure*}

\section{Related Work}

Predicting a sequence of future center positions at a coarse-grained level~\cite{kothari2020human, saadatnejad2022sattack, bahari2022sattack} or a sequence of bounding boxes~\cite{bouhsain2020pedestrian,saadatnejad2022pedestrian} have been extensively studied in the literature. However, in this work, we focus on a more fine-grained prediction, namely 3D pose.
Recurrent Neural Networks (RNNs) have been widely used~\cite{fragkiadaki2015recurrent, jain2016structural, martinez2017human, chao2017forecasting, ghosh2017learning, chiu2019action} as they are capable of capturing the temporal dependencies in sequential data, and later networks with only feed-forward networks were introduced~\cite{li2018convolutional}.
Subsequently, Graph Convolutional Networks (GCNs) were proposed to better capture the spatial dependencies of body poses~\cite{mao2019ltd,cui20ldrgcn,mao2020history,liu2021mpt}. 
Separating temporal and spatial convolution blocks~\cite{ma2022progressively}, and trainable adjacency matrices~\cite{sofianos2021stsgcn,zhong2022stgagcn} are among other proposed ideas.
Attention-based approaches have recently gained interest for modeling human motion~\cite{martinezgonzalez2021posetransformers,petrovich2021actionconditioned} and showed a huge improvement with spatio-temporal self-attention module~\cite{mao2020history}.
Our proposed model also incorporates attention. 
While various works have employed context information~\cite{cao2020long,hassan2021stichastic,corona2020context}, social interactions~\cite{adeli2020socially} or action classes~\cite{aksan2020motiontransformer, cai2021unified} as conditions, this paper focuses on conditioning solely on the observation sequences.

% generative models
Deterministic models~\cite{mao2020history,ma2022progressively} offer satisfactory prediction accuracy, yet they lack the ability to generate diverse and multi-modal outputs compared to stochastic models~\cite{yuan2020dlow, aliakbarian2020stochastic, aliakbarian2021contextually, salzmann2022motron, ma2022multiobjective, mao2021generating,xu22stars}. 
In this category, Variational AutoEncoders (VAEs) have been widely adopted due to their strength in representation learning~\cite{parsaeifard2021decoupled,yuan2020dlow, aliakbarian2020stochastic, aliakbarian2021contextually}.
% diffusion models
Generative models, particularly diffusion models, have been recently utilized to model data distributions with remarkable results in image synthesis~\cite{dhariwal2021diffbeatgan,saadatnejad2021semdisc}, image repainting~\cite{lugmayr2022repaint} and text-to-image generation~\cite{saharia2022imagen, rombach2022ldm}.
Recently, they have been used for time-series imputation~\cite{tashiro2021csdi}, i.e., filling in missing elements.
However, it was not explored for human motion.
To the best of our knowledge, we are the first to propose a diffusion model for human pose prediction, which outperforms both stochastic and deterministic models.

% noisy data
Previous models perform poorly with partial noisy observations.
A multi-task learning approach has been recently suggested in~\cite{cui2021incomplete} to address this issue, by implicitly disregarding noise in the data.
We provide detailed comparisons with~\cite{cui2021incomplete}, and show that explicitly denoising the input leads to a generalizable solution, and that our temporally-cascaded diffusion blocks better capture the spatio-temporal relationships in the poses.
Furthermore, we present a generic framework that can be used to improve any existing state-of-the-art model in a black-box manner.

\section{Method}
In this section, we first describe the notations and conditional diffusion blocks, which are the fundamental elements of our model. We then present our model and finally introduce our generic framework.
\subsection{Problem Definition and Notations}
Let $X = [X_{-O+1}, X_{-O+2}, \dots, X_{0}, X_1, \dots, X_P] \in \mathbb{R}^{(O+P) \times J \times 3}$ be a clean complete normalized sequence of human body poses with $J$ joints in $O$ frames of observation and $P$ frames of future. Each joint consists of its 3D cartesian coordinates.
The availability mask is a binary matrix $M \in \{0, 1\}^{(O+P) \times J \times 3}$ where zero determines the parts of the sequence that are not observed due to occlusions or being from future timesteps.
Note that the elements of $M$ corresponding to $P$ future frames are always zero.
With this notation, the observed  sequence $\Tilde{X} = [\Tilde{X}_{-O+1}, \Tilde{X}_{-O+2}, \dots, \Tilde{X}_{0}, \Tilde{X}_{1}, \dots, \Tilde{X}_{P}] $ is derived by applying the element-wise product of $M$ into $X$ and adding a Gaussian noise $\epsilon \sim \mathcal{N}(0,\,I)$ in non-masked area $\Tilde{X} = M \odot X + (1-M) \epsilon$.
The model predicts $\hat{X} = [\hat{X}_{-O+1}, \hat{X}_{-O+2}, \dots, \hat{X}_{0}, \hat{X}_{1}, \dots, \hat{X}_{P}] $ and the objective is lowering $|\hat{X} - X| \odot (1-M)$ given $\Tilde{X}$.

\begin{figure*}
    \centering
    \vspace{2pt}
    \includegraphics[width=\textwidth]{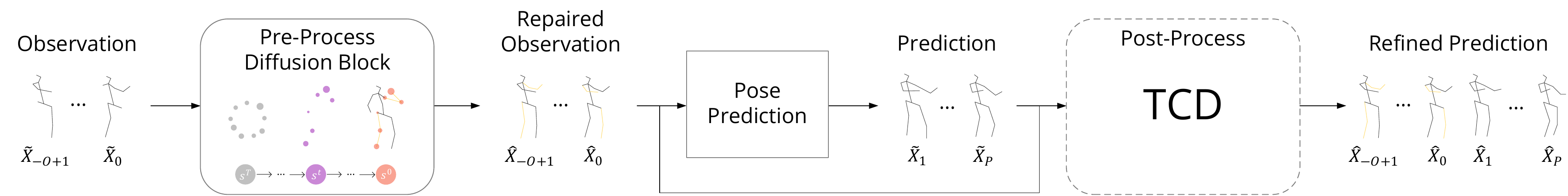}
    \caption{An illustration of the pre-processing and post-processing framework. The pre-process diffusion block denoises the noisy observation sequence. The repaired observation is then given to a frozen predictor. The output of the predictor model is passed to TCD to perform the post-processing step and refine its predictions.}
    \label{fig:pre-post-process}
    \vspace{-2pt}
\end{figure*}

\subsection{Conditional Diffusion Blocks}
We propose a conditional diffusion block, inspired by~\cite{tashiro2021csdi}, which contains multiple residual layers. Each layer consists of two consecutive transformers with the same input and output shapes.
The first (temporal) transformer is responsible for modeling the temporal behavior of data. Its output is then fed to the second (spatial) transformer to attend to the body pose within each frame.

At training time, a Gaussian noise with zero mean and pre-defined variance is added to the input pose sequence $s^0$ to make a noisier version $s^1$. This process is repeated for $T$ steps such that the output $s^T$ will be close to a pure Gaussian noise in the non-masked area:
\begin{equation}
    q(s^{t} | s^{t-1})
    = M \odot s^{t-1} + (1-M) \odot \mathcal{N}(s^t; \sqrt{1-\beta^t} s^{t-1}, \beta^t \mathbf{I}),
    \label{eq:singlestep}
\end{equation}
where $q$ denotes the forward process, and $\beta^t$ is the variance of the noise in step $t$, determined using a scheduler.
We use the cosine noise scheduler in our formulations, which was first introduced in~\cite{nichol2021improved}:
\begin{equation}
    \beta^t = 1 - \frac{f(t)}{f(t-1)}, \quad
    f(t) = \cos^2\biggl({\frac{t/T + c}{1 + c} \cdot \frac{\pi}{2}}\biggr),
\end{equation}
where $c$ is a small offset and is set to $0.008$ empirically.
The cosine noise scheduler provides a smoother decrease in input quality than other popular schedulers, such as quadratic and linear~\cite{nichol2021improved}, enabling more accurate learning of step noise variances in our problem.
The network learns to reverse the diffusion process and retrieve the clean sequence by predicting the cumulative noise that is added
to $s^t$ as described in DDPM~\cite{ho2020ddpm}.

At inference time, the model begins with an incomplete and noisy input sequence $s^T$, where Gaussian noise is put in the non-masked area and observed data in the masked area. Subsequently, the model iteratively predicts the poses $s^{T-1}, \dots, s^0$ through an iterative process by subtracting the additive noise learned during training from the output of the preceding step, until a clean output approximating the ground truth is obtained.

\subsection{Temporal Cascaded Diffusion (TCD)}
We illustrate our main model, which consists of a short-term and a long-term diffusion blocks, in \Cref{fig:model}.
The short-term block takes $\Tilde{X}$ as input and predicts the first $K$ frames of the future $[\hat{X}_1 \dots \hat{X}_K]$, along with the observation frames $[\hat{X}_{-O+1} \dots \hat{X}_{0}]$.
The long-term block is tasked with predicting the remaining frames of the future $[\hat{X}_{K+1} \dots \hat{X}_{P}]$, utilizing both the observation and the output of the short-term block.
Note that during training, both blocks are trained using ground-truth input; however, at inference time, the average of five samples of the short-term block is supplied to the long-term block.

Cascading two diffusion models improves overall and particularly long-term forecasting due to the division of the complex task.
In other words, the short-term prediction block focuses on predicting a limited number of frames, and thanks to its accurate short-term predictions, the long-term prediction block acquires more data, thus allowing it to focus its capacity on longer horizons.

\subsection{Pre-processing and Post-processing}
Given a frozen pose prediction model, we can enhance its performance through pre-processing by repairing its input sequence, and through post-processing by refining its outputs. This framework is illustrated in \Cref{fig:pre-post-process}.

\paragraph{Pre-Processing}
Since most of the existing pose prediction models are unable to handle noisy observations, we present a simpler version of our model that serves as a pre-processing step for denoising the observations only.
This module takes the noisy observation sequence $[\Tilde{X}_{-O+1}, \Tilde{X}_{-O+2}, \dots, \Tilde{X}_{0}]$ as input and outputs a repaired sequence $[\hat{X}_{-O+1}, \hat{X}_{-O+2}, \dots, \hat{X}_{0}]$. The architecture of this model is similar to TCD, yet predicting within a single stage, with both the input and output sequences containing $O$ frames. Our precise repair strategies allow any pose prediction models trained on complete datasets to predict reasonable poses in noisy input conditions.

\paragraph{Post-Processing}
Furthermore, we want to improve the prediction results of existing models.
We feed the results of any black-box pose prediction model $[\Tilde{X}_{1}, \dots, \Tilde{X}_{P}]$ concatenated with repaired observation $[\hat{X}_{-O+1}, \hat{X}_{-O+2}, \dots, \hat{X}_{0}$] as the input to our TCD and retrain it to predict better. The initial prediction acts as the starting point that is gradually shifted toward the real distribution by our post-processing.

\begin{table*}[!t]
	\centering
        \vspace{2pt}
    	\begin{tabular}{lcccccc}
    	    \toprule
    		& \multicolumn{4}{c}{Human3.6M~\cite{h36m}} & \multicolumn{2}{c}{HumanEva-I~\cite{sigal2010humaneva}} \\
    		\cmidrule(lr){2-5} \cmidrule(lr){6-7}
    		Model & ADE $\downarrow$ & FDE $\downarrow$ & MMADE $\downarrow$ & MMFDE $\downarrow$ & ADE $\downarrow$ & FDE $\downarrow$ \\
    		\midrule
    		Pose-Knows~\cite{walker2017pose} & 461 & 560 & 522 & 569 & 269 & 296 \\
    		MT-VAE~\cite{yan2018mt}  & 457 & 595 & 716 & 883 & 345 & 403 \\ 
    		HP-GAN~\cite{Barsoum2018HPGANP3} & 858 & 867 & 847 & 858 & 772 & 749 \\
    		BoM~\cite{Bhattacharyya2018AccurateAD} & 448 & 533 & 514 & 544 & 271 & 279 \\ 
    		GMVAE~\cite{DBLP:journals/corr/DilokthanakulMG16} & 461 & 555 & 524 & 566 & 305 & 345 \\ 
    		DeLiGAN~\cite{Gurumurthy2017DeLiGANGA} & 483 & 534 & 520 & 545 & 306 & 322 \\ 
    		DSF~\cite{yuan2019diverse}  & 493 & 592 & 550 & 599 & 273 & 290 \\ 
    		DLow~\cite{yuan2020dlow}  & 425 & 518 & 495 & 531 & 251 & 268 \\ 
    		Motron~\cite{salzmann2022motron}  & 375 & 488 & -- & -- & -- & -- \\
    		Multi-Objective~\cite{ma2022multiobjective}  & 414 & 516 & -- & -- & 228 & 236 \\
    		GSPS~\cite{mao2021generating} & 389 & 496 & 476 & 525 & 233 & 244 \\
            STARS~\cite{xu22stars} & 358 & 445 & 442 & 471 & 217 & 241 \\
    		TCD (ours) & \textbf{356} & \textbf{396} & \textbf{463} & \textbf{445} & \textbf{199} & \textbf{215} \\
    		\bottomrule
    	\end{tabular}
\caption{Comparison with stochastic models on Human3.6M~\cite{h36m} Setting-A and HumanEva-I~\cite{sigal2010humaneva} at a horizon of 2s.}
\vspace{-2pt}
\label{tab:h36_sto}
\end{table*}

\section{Experiments}

\subsection{Experimental Setup}

\subsubsection{Datasets}

We evaluate the performance of all approaches on four widely-used 3D pose prediction datasets:

\textbf{Human3.6M}~\cite{h36m} is the largest benchmark dataset for human motion analysis, comprising 3.6 million body poses. It consists of 15 complex action categories, each performed by seven actors individually.
The training set comprises five subjects, and the validation and test sets comprise two different subjects.
We train our models on all action classes concurrently.
The original 3D pose skeletons in the dataset consist of 32 joints, but different subsets of joints have been used in previous works to represent human poses.
To ensure a fair and comprehensive comparison, we consider three different settings for the dataset as follows:
\begin{itemize}
    \item \textbf{Setting-A}: 25 observation frames, 100 prediction frames at 50 frames per second (fps), with the subset of 17 joints to represent the human pose; % similar to~\cite{ma2022multiobjective}
    \item \textbf{Setting-B}: 50 observation frames, 25 prediction frames down-sampled to 25 fps, with the subset of 22 joints to represent the human pose; %similar to~\cite{mao2020history}
    \item \textbf{Setting-C}: 25 observation frames, 25 prediction frames down-sampled to 25 fps, with the subset of 17 joints to represent the human pose. %similar to~\cite{cui2021incomplete}
\end{itemize}

\textbf{AMASS} (Archive of Motion capture As Surface Shapes)~\cite{amass2019} is a recently published human motion dataset that combines 18 motion capture datasets, totaling 13,944 motion sequences from 460 subjects performing various actions.
We use 50 observation frames down-sampled to 25 fps with 18 joints, as in previous studies.

\textbf{3DPW} (3D Poses in the Wild)~\cite{3dpw} is the first dataset with accurate 3D poses in the wild. It contains 60 video sequences and each pose is described with an 18-joint skeleton, similar to the AMASS dataset. We use the official instructions to obtain training, validation, and test sets.

\textbf{HumanEva-I}~\cite{sigal2010humaneva} includes three subjects captured at 60 fps. Each person has 15 body joints. We remove the global translation and use the official train/test split of the dataset. The prediction horizon is 60 frames (1 second) given 15 observed frames (0.25 seconds), similar to~\cite{mao2021generating}.

\subsubsection{Other Implementation Details}
We train our models using the Adam optimizer~\cite{adam}, with a batch size of $32$ and a learning rate of $0.001$. The learning rate is decayed by a factor of $0.1$ at $75\%$ and $90\%$ of the total epochs.
Our model consists of $12$ layers of residual blocks and $50$ diffusion steps by default.
In TCD, the length of short-term prediction $K$ is set to $20\%$ of the total prediction length $P$.
Each transformer has $64$ channels and $8$ attention heads.

\begin{table*}[!t]
    \centering
    \vspace{2pt}
    \begin{tabular}{lcccccc}
        \toprule
        Model & 80ms & 320ms & 560ms & 720ms & 880ms & 1000ms \\
        \midrule
        Zero-Vel & 23.8 & 76.0 & 107.4 & 121.6 & 131.6 & 136.6 \\
        Res. Sup.~\cite{martinez2017human} & 25.0 & 77.0 & 106.3 & 119.4 & 130.0 & 136.6 \\ 
        ConvSeq2Seq~\cite{li2018convolutional} & 16.6 & 61.4 & 90.7 & 104.7 & 116.7 & 124.2 \\ 
        LTD-50-25~\cite{mao2019ltd} & 12.2 & 50.7 & 79.6 & 93.6 & 105.2 & 112.4\\
        HRI~\cite{mao2020history} & 10.4 & 47.1 & 77.3 & 91.8 & 104.1 & 112.1\\
        PGBIG~\cite{ma2022progressively} & 10.3 & \textbf{46.6} & 76.3 & 90.9 & 102.6 & 110.0\\
        TCD (ours) & \textbf{9.9} & 48.8 & \textbf{73.7} & \textbf{84.0} & \textbf{94.3} & \textbf{103.3}\\
        \midrule
        HRI~\cite{mao2020history} + TCD (ours) & 10.3 & 47.3 & 72.9 & 83.8 & 94.0 & 102.9 \\
        PGBIG~\cite{ma2022progressively} + TCD (ours) & 10.2 & 46.1 & 72.4 & 83.6 & 93.9 & 102.8\\
        \bottomrule
    \end{tabular}
    \caption{Comparison with deterministic models on Human3.6M~\cite{h36m} Setting-B in FDE (mm) at different horizons.}
    \vspace{-2pt}
    \label{tab:h36_det}
\end{table*}

\begin{table*}[!t]
    \centering
    \begin{tabular}{lcccccccc}
        \toprule
    		& \multicolumn{4}{c}{AMASS~\cite{amass2019}} & \multicolumn{4}{c}{3DPW~\cite{3dpw}} \\
    	\cmidrule(lr){2-5} \cmidrule(lr){6-9}
        Model & 560ms & 720ms & 880ms & 1000ms & 560ms & 720ms & 880ms & 1000ms \\
        \midrule
        Zero-Vel & 130.1 & 135.0 & 127.2 & 119.4 & 93.8 & 100.4 & 102.0 & 101.2 \\
        convSeq2Seq~\cite{li2018convolutional} & 79.0 & 87.0 & 91.5 & 93.5 & 69.4 & 77.0 & 83.6 & 87.8 \\
        LTD-10-25~\cite{mao2019ltd} & 57.2 & 65.7 & 71.3 & 75.2 & 57.9 & 65.8 & 71.5 & 75.5 \\
        HRI~\cite{mao2020history} & 51.7 & 58.6 & 63.4 & 67.2 & 56.0 & 63.6 & 69.7 & 73.7 \\
        TCD (ours) & \textbf{49.8} & \textbf{54.5} & \textbf{60.1} & \textbf{66.7} & \textbf{55.4} & \textbf{61.6} & \textbf{67.9} & \textbf{73.4} \\
        \bottomrule
    \end{tabular}
    \caption{Comparison with deterministic models on AMASS~\cite{amass2019} and 3DPW~\cite{3dpw} in FDE (mm) at long horizons.}
    \label{tab:amass_3dpw}
    \vspace{-5pt}
\end{table*}

\subsubsection{Evaluation Metrics}

We measure the Displacement Error (DE), in millimeters (mm), over all joints in a frame.
Then, we report the Average Displacement Error (ADE), which is the average DE across all prediction frames, and/or the Final Displacement Error (FDE), which is the DE in the final predicted frame.
We also report the multi-modal versions of ADE (MMADE) and FDE (MMFDE), following~\cite{mao2021generating}.
We additionally report ADE for the missing joints of the observation frames in the repairing task (r-ADE).

\subsection{Baselines}
We compare our model with several recent methods, including stochastic ~\cite{mao2021generating,yuan2020dlow,salzmann2022motron,ma2022multiobjective,xu22stars} and deterministic approaches~\cite{martinez2017human, li2018convolutional, mao2019ltd, mao2020history, ma2022progressively, sofianos2021stsgcn, zhong2022stgagcn} when possible. Note that some methods are not open-source and have different settings than ours.
We also include \textit{Zero-Vel} as a competitive baseline. \textit{Zero-Vel} is a simple model that predicts the last observed pose for all future frames.

\subsection{Comparisons with the State of the Art}

We separate our experiments into three different settings: we first compare to other stochastic approaches, then to deterministic ones, and finally evaluate on noisy scenarios, with missing or noisy observation data.

\subsubsection{Comparisons with Stochastic Approaches}
\label{sec:stochastic}
We evaluate our model on two datasets, Human3.6M~\cite{h36m} Setting-A and HumanEva-I~\cite{sigal2010humaneva}, and compare it with other stochastic approaches in \Cref{tab:h36_sto}.
Each model is sampled $50$ times given each observation sequence.
TCD (ours) clearly outperforms all previous works in terms of accuracy of the best sample (as measured by ADE and FDE) and multiple samples (as measured by MMADE and MMFDE).

\subsubsection{Comparisons with Deterministic Approaches}
\label{sec:det}
We then compare our model to deterministic approaches on Human3.6M~\cite{h36m} Setting-B, tabulated in \Cref{tab:h36_det}.
To compare with deterministic models, our model is sampled five times, and the best sample is considered.
Our proposed model surpassed previous works in the short-term and with a marked margin in the long-term, thanks to our two-stage prediction strategy. The detailed results of our model's performance on all categories of Human3.6M, along with comparisons with models that are not reported in standard settings, can be found in the appendix.
We have also included the results of two previous state-of-the-art models that have been post-processed by our generic framework at the bottom of \Cref{tab:h36_det}. Note that as the input data is complete, we only add post-processing (TCD) to their outputs. The improvements from our framework are non-negligible and can even beat our original model.
Our two-stage prediction reveals a more pronounced benefit for longer horizons, which suggests that starting with a better initial guess can better shift the pose sequence toward the real distribution.

Substantial long-term improvement can be observed in AMASS~\cite{amass2019} and 3DPW~\cite{3dpw} as well. Similar to previous works, we train our model on AMASS and measure the FDE on both datasets. The comparison with models reporting in this setting is in \Cref{tab:amass_3dpw}. Note that for faster training, $K=0$ was considered in this experiment.

\begin{figure}
    \centering
    \includegraphics[width=\columnwidth]{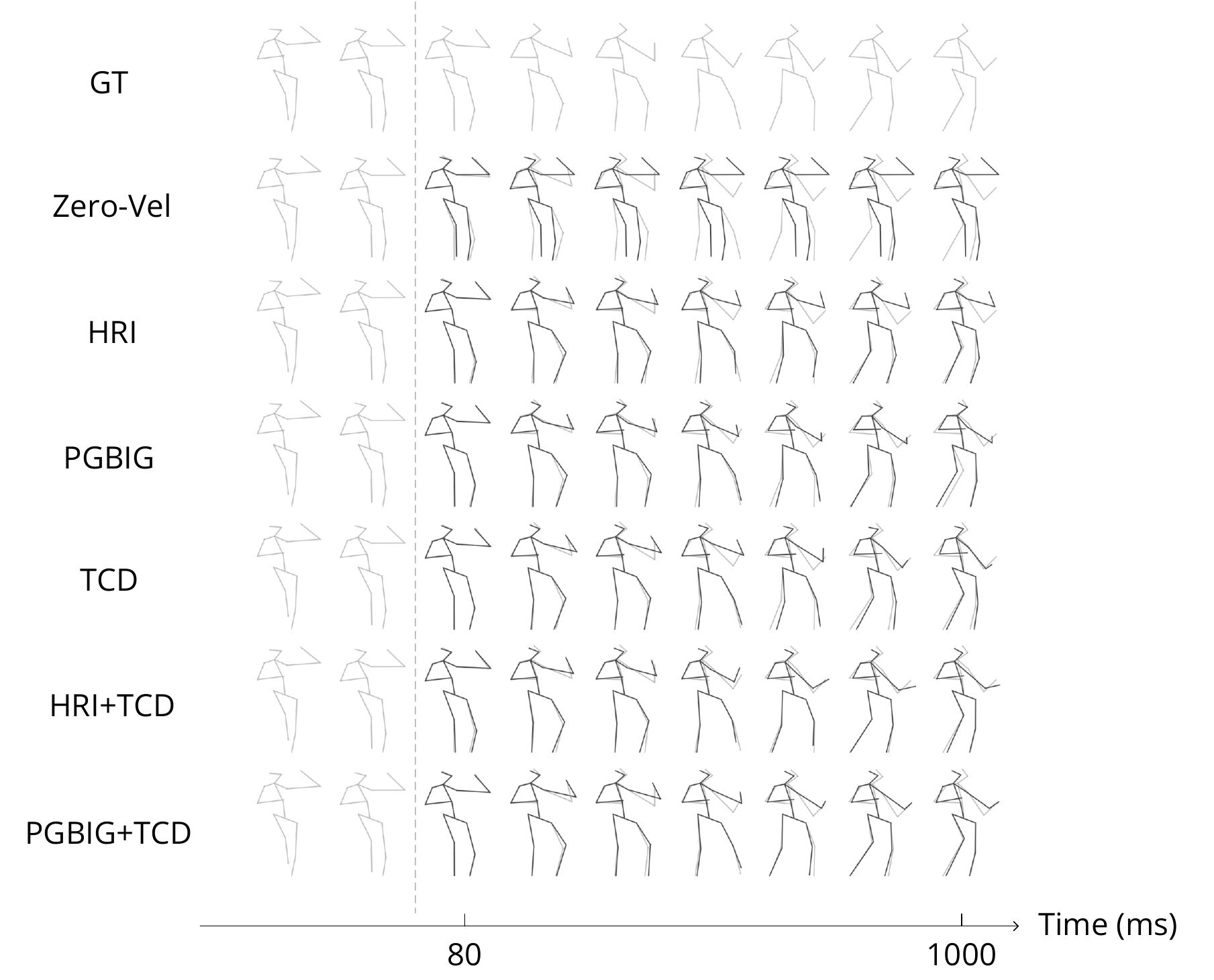}
    \caption{Qualitative results on Human3.6M~\cite{h36m} Setting-B. The left part of each row shows the input observation, while the right part displays the predicted poses superimposed on the ground truth.}
    \label{fig:h36}
    \vspace{-10pt}
\end{figure}

Qualitative results on Human3.6M are shown in \Cref{fig:h36}.
Predictions from our model are displayed along with predictions from several baselines and are superimposed on the ground-truth poses for direct comparison.
Our model has successfully learned the data distribution, resulting in accurate and realistic poses; for instance, the hand movement is natural when the feet move while HRI has fixed hands and PGBIG has a momentum that avoids large hand movements.
Moreover, post-processing can be used to further refine the predicted pose and shift it toward the ground truth.

\subsubsection{Comparisons on Noisy Observation Data}

We now examine the performance of models in the realistic scenario of noisy observations, since occlusions and noise are commonly seen in practice.
To simulate occlusions, we remove $40\%$ of the left arm and right leg from the observations of Human3.6M Setting-B, both during training and evaluation. The results in the top half of \Cref{tab:h36_inc_ourprep} show that the state-of-the-art models perform inadequately when the observation is noisy, whereas our model achieves results close to those of the clean input observation.
Our pre-processing module repairs the observation sequences before feeding to the state-of-the-art models and Zero-Vel, resulting in significant improvements in forecasting performance.
MT-GCN~\cite{cui2021incomplete} was designed to provide accurate predictions in incomplete observations. We compared our model to it and some other prior models and present the results on Human3.6M Setting-C in the first column of \Cref{tab:inc_pattern}. Our model achieved a remarkable improvement of 33.2mm in FDE at 1s horizon (30\% improvement) over MT-GCN. It should be noted that the models in the upper part of the table received repaired sequences using MT-GCN's own preprocessing, while the rest received noisy sequences.

\begin{table}[!t]
    \vspace{2pt}
    \centering
    \resizebox{\linewidth}{!}{
        \begin{tabular}{lcccccc}
            \toprule
            Model & 80ms & 320ms & 560ms & 720ms & 880ms & 1000ms \\
            \midrule
            Zero-Vel & 84.9 & 138.2 & 169.9 & 184.2 & 193.7 & 198.2\\
            HRI~\cite{mao2020history} & 65.2 & 104.5 & 130.0 & 141.6 & 151.1 & 157.1\\
            PGBIG~\cite{ma2022progressively} & 67.0 & 107.1 & 132.1 & 143.5 & 152.9 & 158.8\\
            TCD (ours) & \textbf{11.2} & \textbf{51.3} & \textbf{75.4} & \textbf{85.4} & \textbf{95.4} & \textbf{104.5}\\
            \midrule
            Pre(ours) + Zero-Vel & 24.1 & 76.3 & 107.6 & 121.7 & 131.7 & 136.7\\
            Pre(ours) + HRI~\cite{mao2020history} & 11.4 & 48.6 & 78.3 & 92.7 & 105.0 & 112.8\\
            Pre(ours) + PGBIG~\cite{ma2022progressively} & 11.1 & \textbf{47.9} & 77.2 & 91.7 & 103.5 & 110.8\\
            Pre(ours) + TCD (ours) & \textbf{10.8} & 49.9 & \textbf{74.4} & \textbf{84.9} & \textbf{95.1} & \textbf{104.2}\\
            \bottomrule
        \end{tabular}
    }
    \caption{Comparison on noisy observation data and pre-processed observation data (Pre(ours)+) on Human3.6M~\cite{h36m} Setting-B in FDE (mm) at different horizons.}
    \vspace{-2pt}
    \label{tab:h36_inc_ourprep}
\end{table}

\begin{table}[!t]
    \centering
    \resizebox{\linewidth}{!}{
        \begin{tabular}{lcccc}
            \toprule
            Model & \makecell{Random\\Leg, Arm\\Occlusions} & \makecell{Structured\\Joint\\Occlusions} & \makecell{Missing\\Frames} & \makecell{Gaussian Noise\\$\sigma=25 \ $ $\sigma=50$} \\
            \midrule
            R+TrajGCN~\cite{mao2019ltd} & 121.1 & 131.5 & -- & 127.1  \quad 135.0 \\
            R+LDRGCN~\cite{cui20ldrgcn} & 118.7 & 127.1 & -- & 126.4  \quad 133.6 \\
            R+DMGCN~\cite{li2020dynamic} & 117.6 & 126.5 & -- & 124.4 \quad 132.7\\
            R+STMIGAN~\cite{hernandez2019human} & 129.5 & 128.2 & -- & -- \quad\quad\quad --\\
            \midrule
            MT-GCN~\cite{cui2021incomplete} & 110.7 & 114.5 & 122.0 & 114.3 \quad 119.7 \\
            TCD (ours) & \textbf{77.5} & \textbf{77.2} & \textbf{80.5} & \textbf{81.9} \quad \, \textbf{84.9} \\
            \bottomrule
        \end{tabular}
    }
    \caption{Comparison on noisy observation data on Human3.6M~\cite{h36m} Setting-C in FDE (mm) at a horizon of 1s. The upper part of the table contains models that received repaired sequences (R+), while the lower part contains models that received noisy sequences.}
    \label{tab:inc_pattern}
\end{table}

We analyzed the performance of our model in several occlusion patterns masks $M$ that are applied to input data:
\begin{itemize}
    \item Random Leg, Arm Occlusions: leg and arm joints are randomly occluded with a probability of $40\%$;
    \item Structured Joint Occlusions: $40\%$ of the right leg joints for consecutive frames are missing;
    \item Missing Frames: $20\%$ of the consecutive frames are missing;
    \item Gaussian Noise: Gaussian noise with a standard deviation of $\sigma=25$ or $\sigma=50$ is added to the coordinates of the joints, and 50\% of the leg joints are randomly occluded.
\end{itemize}
The results of training and evaluating our model on these observation patterns, in FDE at a prediction horizon of $1$ second on Human3.6M Setting-C, are presented in \Cref{tab:inc_pattern}.
Our model outperformed previous works in different patterns of occlusions and noises in input that can occur in the real world. Furthermore, we observed that missing $5$ consecutive frames is more challenging than missing a part of the body in $10$ consecutive frames, as the network can recover the latter with spatial information.

\begin{table}[!t]
    \centering
    \resizebox{\linewidth}{!}{
    \begin{tabular}{lcccc}
        \toprule
        & \multicolumn{4}{c}{Train and Test Missing Ratio} \\
        \cmidrule(lr){2-5}
        Model & 10\% & 20\% & 30\% & 40\% \\
        \midrule
        MT-GCN~\cite{cui2021incomplete} & 109.4 / 8.6 & 110.5 / 13.7 & 112.3 / 18.7 & 114.4 / 24.5\\
        TCD (ours) & \textbf{77.1 / 2.2} & \textbf{77.2 / 2.3} & \textbf{77.6 / 2.6} & \textbf{ 79.1 / 2.9} \\
        \bottomrule
    \end{tabular}
    }
    \caption{Results of motion prediction and sequence repairing on Human3.6M~\cite{h36m} Setting-C with varying amounts of randomly occluded joints in input data in FDE (mm) at a horizon of 1s / r-ADE (mm) of missing elements.}
    \vspace{-3pt}
    \label{tab:h36_inc_dif}
\end{table}

To have a thorough comparison with MT-GCN, we trained four models by varying the percentage of joints randomly removed from the pose observation sequence. 
The performance of sequence repairing (r-ADE of the occluded observation sequence) and motion prediction (FDE at 1-second horizon) is presented in \Cref{tab:h36_inc_dif}.
Our model exhibited a negligible error of $2.9$mm in repairing with up to $40\%$ of all joints missing, whereas MT-GCN exhibited an error of $24.5$mm. Indeed, our model achieved more than 31\% lower FDE compared to MT-GCN in forecasting.

\subsection{Ablations Studies}
Here, we investigate different design choices of the network and report ADE  on Human3.6M~\cite{h36m} Setting-B. 
For faster training, only a fifth of the dataset was utilized in this section. 
The full model yielded an ADE of $63.3$mm. When predicting in one stage, without any subdivisions, the ADE increased to $65.5$mm due to erroneous predictions in longer time frames. Conversely, when predicting in three stages, i.e., $20\%$, $20\%$, and $60\%$, the performance dropped to $66.9$mm, as cascading multiple stochastic processes leads to either random outcomes or a lack of diversity. This illustrates the efficacy of two-stage prediction. 
Another important factor is the length of short-term prediction. In our experiments, a prediction of $P=25$ frames was made with $K=5$. A lower $K=2$ reduced the benefits of two-stage prediction (ADE of $65.1$mm). On the other hand, a higher $K=10$ made short-term prediction more difficult, leading to an increased ADE of $66.6$mm.

We tested a quadratic scheduler instead of our cosine scheduler and it increased ADE by $1$mm.
Our full model employed 12 residual layers in its diffusion blocks; however, decreasing this number to 4 resulted in a decrease in performance by $3$mm. We refrained from utilizing more than 12 residual layers due to the considerable negative influence on the sampling time.
Moreover, we conducted several experiments on the architecture of the transformers and found that spatial transformer and time transformer both facilitated the learning of spatio-temporal features of the pose sequence. Eliminating either of these resulted in an ADE of $74.5$mm and $261.1$mm, respectively.

\section{Conclusion}
In this work, we proposed a denoising diffusion model for 3D human pose prediction suitable for noisy input observations occurring in the wild.
Our model predicted future poses in two stages (short-term and long-term) to better capture human motion dynamics, achieved superior performance compared to the state-of-the-art on four datasets, including both clean and noisy input settings.
We then leveraged it to create a generic framework that is easily applicable to any existing predictor in a black box manner in two steps: pre-processing to repair the observations and post-processing to refine the predicted poses. We have applied it to several previous predictors and enhanced their predictions. The high computational complexity of diffusion models is a well-known challenge, and future studies may explore ways to accelerate the model's performance without sacrificing accuracy.

%===============================================================================
\section*{ACKNOWLEDGMENT}
The authors would like to thank Mohammadhossein Bahari and Bastien Van Delft for their helpful comments.
This project has received funding from the European Union's Horizon 2020 research and innovation programme under the Marie Sklodowska-Curie grant agreement No 754354, and SNSF Sinergia Fund.

%===============================================================================

% \clearpage

{\small
\bibliographystyle{ieee_fullname}
\bibliography{references}
}
\newpage

\section*{APPENDIX}
Here, we extend our comparisons in \Cref{sec:det}:

\begin{enumerate}
    \item
    We compared our model's performance with the models that reported ADE in Table \ref{tab:h36_det2}. Our setting was changed to predict 25 frames given 10 observation frames on the Human3.6M dataset down-sampled to 25 fps with the subset of 22 joints (Setting-D), following the settings of~\cite{zhong2022stgagcn}. The evaluation of the models was conducted on all actions except walking together. Our model outperforms those GCN-based models.
    \begin{table}[!h]
    \centering
    \begin{tabular}{lcccccc}
        \toprule
        Model & 80ms & 160ms & 320ms & 400ms & 560ms & 1000ms \\
        \midrule
        Zero-Vel & 18.1 & 28.7 & 46.9 & 54.6 & 67.7 & 93.3 \\
        STSGCN~\cite{sofianos2021stsgcn} & 10.2 & 17.3 & 33.5 & 38.9 & 51.7 & 77.3\\
        GAGCN~\cite{zhong2022stgagcn} & 10.1 & 16.9 & 32.5 & 38.5 & 50.0 & 72.9\\
        TCD (ours) & \textbf{7.4} & \textbf{14.0} & \textbf{27.7} & \textbf{33.9} & \textbf{44.7} & \textbf{66.5}\\
        \bottomrule
    \end{tabular}
    \caption{Comparison with deterministic models on Human3.6M~\cite{h36m} Setting-D in ADE (mm) at different prediction horizons.}
    \label{tab:h36_det2}
    \end{table}
    
    \item
    We conducted another experiment to compare our model's performance with others that reported their results on Human3.6M Setting-E, as shown in Table \ref{tab:h36_det3}. In this setting, 25 frames are predicted given 10 observation frames down-sampled to 25 fps with the subset of 17 joints. Our model outperformed others, particularly in longer horizons.
    \begin{table}[!h]
    \centering
    \begin{tabular}{lcccccc}
        \toprule
        Model & 80ms & 160ms & 320ms & 400ms & 560ms & 1000ms \\
        \midrule
        Zero-Vel & 17.1 & 31.9 & 54.8 & 63.8 & 78.3 & 100.0 \\
        LDRGCN~\cite{cui20ldrgcn} & 10.7 & 22.5 & 45.1 & 55.8 & -- & 97.8 \\
        MPT~\cite{liu2021mpt} & 8.3 & 18.8 & 39.0 & 47.9 & 65.3 & 96.4   \\
        TCD (ours) & \textbf{8.3} & \textbf{18.8} & \textbf{37.8} & \textbf{44.9} & \textbf{55.9} & \textbf{76.9}\\
        \bottomrule
    \end{tabular}
    \caption{Comparison with deterministic models on Human3.6M~\cite{h36m} Setting-E in FDE (mm) at different prediction horizons.}
    \label{tab:h36_det3}
    \end{table}
    
    \item
    In \Cref{tab:h36_det}, we compared the performance of different models on Human3.6M \cite{h36m} Setting-B. The detailed results on all categories are reported in \Cref{tab:h36_det_details}. 
    We observe that in almost all categories, ours beats previous models.

\begin{table*}[!t]
	\resizebox{\textwidth}{!}{
		\begin{tabular}{c|cccccc|cccccc|cccccc|cccccc}
			\toprule
			Scenarios & \multicolumn{6}{c|}{Walking} & \multicolumn{6}{c|}{Eating} & \multicolumn{6}{c|}{Smoking} & \multicolumn{6}{c}{Discussion} \\ \hline
			Model & 80ms & 320ms & 560ms & 720ms & 880ms & 1000ms & 80ms & 320ms & 560ms & 720ms & 880ms & 1000ms & 80ms & 320ms & 560ms & 720ms & 880ms & 1000ms & 80ms & 320ms & 560ms & 720ms & 880ms & 1000ms \\ \hline
			Zero-Vel & 33.9 & 109.8 & 145.9 & 154.4 & 150.7 & 140.2 & 16.5 & 55.3 & 81.3 & 94.4 & 100.7 & 102.1 & 17.3 & 57.1 & 80.3 & 91.4 & 98.1 & 101.1 & 24.5 & 76.8 & 108.7 & 123.5 & 131.5 & 135.3 \\
			Res. Sup. & 23.2 & 61.0 & 71.6 & 72.5 & 76.0 & 79.1 & 16.8 & 53.5 & 74.9 & 85.9 & 93.8 & 98.0 & 18.9 & 57.5 & 78.1 & 88.6 & 96.6 & 102.1 & 25.7 & 80.0 & 109.5 & 122.0 & 128.6 & 131.8 \\
			convSeq2Seq & 17.7 & 56.3 & 72.2 & 77.2 & 80.9 & 82.3 & 11.0 & 40.7 & 61.3 & 72.8 & 81.8 & 87.1 & 11.6 & 41.3 & 60.0 & 69.4 & 77.2 & 81.7 & 17.1 & 64.8 & 98.1 & 112.9 & 123.0 & 129.3 \\
			LTD & 12.3 & 39.4 & 50.7 & 54.4 & 57.4 & 60.3 & 7.8 & 31.3 & 51.5 & 62.6 & 71.3 & 75.8 & 8.2 & 32.8 & 50.5 & 59.3 & 67.1 & 72.1 & 11.9 & 55.1 & 88.9 & 103.9 & 113.6 & 118.5 \\
			HRI & \underline{10.0} & \textbf{34.2} & \underline{47.4} & \underline{52.1} & \underline{55.5} & \underline{58.1} & 6.4 & \textbf{28.7} & 50.0 & 61.4 & 70.6 & 75.7 & \underline{7.0} & \underline{29.9} & \textbf{47.6} & \underline{56.6} & \underline{64.4} & \underline{69.5} & 10.2 & \underline{52.1} & 86.6 & \underline{102.2} & 113.2 & 119.8 \\
			PGBIG & 10.6 & 36.6 & 49.1 & 53.0 & 56.0 & 58.6 & \underline{6.3} & \textbf{28.7} & \underline{49.2} & \underline{60.4} & \underline{68.9} & \underline{73.9} & 7.1 & \textbf{30.1} & \underline{49.2} & 58.9 & 66.4 & 71.2 & \underline{9.9} & \textbf{50.9} & \underline{86.2} & 102.3 & \underline{112.8} & \underline{118.4} \\
			Ours & \textbf{9.9} & \underline{35.7} & \textbf{44.1} & \textbf{46.2} & \textbf{49.8} & \textbf{53.6} & \textbf{6.1} & \underline{29.0} & \textbf{44.5} & \textbf{52.0} & \textbf{59.2} & \textbf{65.1} & \textbf{6.6} & 31.4 & \textbf{47.6} & \textbf{55.2} & \textbf{62.4} & \textbf{68.1} & \textbf{9.6} & 54.9 & \textbf{85.7} & \textbf{96.2} & \textbf{103.6} & \textbf{110.9} \\ \hline
			Scenarios & \multicolumn{6}{c|}{Directions} & \multicolumn{6}{c|}{Greeting} & \multicolumn{6}{c|}{Phoning} & \multicolumn{6}{c}{Posing} \\ \hline
			Model & 80ms & 320ms & 560ms & 720ms & 880ms & 1000ms & 80ms & 320ms & 560ms & 720ms & 880ms & 1000ms & 80ms & 320ms & 560ms & 720ms & 880ms & 1000ms & 80ms & 320ms & 560ms & 720ms & 880ms & 1000ms \\ \hline
			Zero-Vel & 18.8 & 64.4 & 91.6 & 103.8 & 114.9 & 121.1 & 30.8 & 97.3 & 130.6 & 144.8 & 156.3 & 160.5 & 19.9 & 66.8 & 96.5 & 111.0 & 121.6 & 127.5 & 24.7 & 87.2 & 132.4 & 157.9 & 179.8 & 195.0 \\
			Res. Sup. & 21.6 & 72.1 & 101.1 & 114.5 & 124.5 & 129.5 & 31.2 & 96.3 & 126.1 & 138.8 & 150.3 & 153.9 & 21.1 & 66.0 & 94.0 & 107.7 & 119.1 & 126.4 & 29.3 & 98.3 & 140.3 & 159.8 & 173.2 & 183.2 \\
			convSeq2Seq & 13.5 & 57.6 & 86.6 & 99.8 & 109.9 & 115.8 & 22.0 & 82.0 & 116.9 & 130.7 & 142.7 & 147.3 & 13.5 & 49.9 & 77.1 & 92.1 & 105.5 & 114.0 & 16.9 & 75.7 & 122.5 & 148.8 & 171.8 & 187.4 \\
			LTD & 8.8 & 46.5 & 74.2 & \underline{88.1} & \underline{99.4} & \underline{105.5} & 16.2 & 68.7 & 104.8 & 119.7 & 132.1 & 136.8 & 9.8 & 40.8 & 68.8 & 83.6 & 96.8 & 105.1 & 12.2 & 63.1 & 110.2 & 137.8 & 160.8 & 174.8 \\
			HRI & 7.4 & \underline{44.5} & 73.9 & 88.2 & 100.1 & 106.5 & 13.7 & \underline{63.8} & 101.9 & 118.4 & 132.7 & 138.8 & 8.6 & \underline{39.0} & 67.4 & 82.9 & 96.5 & 105.0 & 10.2 & \underline{58.5} & 107.6 & 136.8 & 161.4 & 178.2 \\
			PGBIG & \underline{7.2} & \textbf{43.5} & \underline{73.1} & 88.8 & 100.5 & 106.1 & \underline{13.4} & \textbf{63.1} & \underline{100.4} & \underline{117.7} & \underline{130.5} & \underline{136.1} & \underline{8.4} & \textbf{38.3} & \underline{66.3} & \underline{82.0} & \underline{95.4} & \underline{103.3} & \underline{9.8} & \textbf{56.5} & \underline{101.5} & \underline{127.8} & \underline{149.9} & \underline{165.3} \\
			Ours & \textbf{7.0} & 46.9 & \textbf{70.6} & \textbf{79.8} & \textbf{90.7} & \textbf{100.3} & \textbf{13.0} & 68.8 & \textbf{98.2} & \textbf{106.2} & \textbf{116.4} & \textbf{126.1} & \textbf{8.0} & 39.6 & \textbf{65.1} & \textbf{77.3} & \textbf{88.8} & \textbf{98.0} & \textbf{9.0} & 59.7 & \textbf{99.5} & \textbf{120.3} & \textbf{138.5} & \textbf{154.1} \\\hline
			Scenarios & \multicolumn{6}{c|}{Purchases} & \multicolumn{6}{c|}{Sitting} & \multicolumn{6}{c|}{Sitting Down} & \multicolumn{6}{c}{Taking Photo} \\ \hline
			Model & 80ms & 320ms & 560ms & 720ms & 880ms & 1000ms & 80ms & 320ms & 560ms & 720ms & 880ms & 1000ms & 80ms & 320ms & 560ms & 720ms & 880ms & 1000ms & 80ms & 320ms & 560ms & 720ms & 880ms & 1000ms \\ \hline
			Zero-Vel & 27.0 & 80.6 & 112.1 & 127.2 & 139.7 & 148.0 &	17.0 & 56.0 & 85.2 & 101.0 & 114.4 & 122.7 & 24.5 & 74.8 & 111.0 & 129.6 & 144.4 & 155.1 & 17.0 & 57.2 & 88.4 & 105.2 & 118.3 & 127.2 \\
			Res. Sup. & 28.7 & 86.9 & 122.1 & 137.2 & 148.0 & 154.0 & 23.8 & 78.0 & 113.7 & 130.5 & 144.4 & 152.6 & 31.7 & 96.7 & 138.8 & 159.0 & 176.1 & 187.4 & 21.9 & 74.0 & 110.6 & 128.9 & 143.7 & 153.9 \\
			convSeq2Seq & 20.3 & 76.5 & 111.3 & 129.1 & 143.1 & 151.5 & 13.5 & 52.0 & 82.4 & 98.8 & 112.4 & 120.7 & 20.7 & 70.4 & 106.5 & 125.1 & 139.8 & 150.3 & 12.7 & 52.1 & 84.4 & 102.4 & 117.7 & 128.1 \\
			LTD & 15.2 & 64.9 & 99.2 & 114.9 & 127.1 & 134.9 & 10.4 & 46.6 & 79.2 & 96.2 & 110.3 & 118.7 & 17.1 & 63.6 & 100.2 & 118.2 & 133.1 & 143.8 & 9.6 & 43.3 & 75.3 & 93.5 & 108.4 & 118.8 \\
			HRI & 13.0 & \underline{60.4} & \underline{95.6} & \underline{110.9} & 125.0 & 134.2 & 9.3 & 44.3 & 76.4 & 93.1 & 107.0 & 115.9 & 14.9 & \underline{59.1} & 97.0 & 116.1 & 132.1 & 143.6 & 8.3 & \underline{40.7} & 72.1 & 90.4 & 105.5 & 115.9 \\
			PGBIG & \underline{12.9} & \textbf{60.1} & \underline{95.6} & 111.1 & \underline{123.1} & \underline{130.6} & \underline{9.0} & \textbf{42.5} & \underline{74.7} & \underline{91.3} & \underline{105.2} & \underline{114.0} & \underline{14.5} & \textbf{58.0} & \underline{95.7} & \underline{114.9} & \underline{130.1} & \underline{140.8} & \textbf{8.1} & \textbf{40.1} & \underline{72.0} & \underline{90.2} & \underline{105.2} & \underline{115.4} \\
			Ours & \textbf{12.1} & 60.9 & \textbf{88.9} & \textbf{100.0} & \textbf{112.3} & \textbf{123.3} & \textbf{8.7} & \underline{43.8} & \textbf{71.3} & \textbf{85.2} & \textbf{98.5} & \textbf{108.1} & \textbf{14.1} & 61.3 & \textbf{94.2} & \textbf{110.3} & \textbf{124.6} & \textbf{135.7} & \underline{8.2} & 42.6 & \textbf{70.5} & \textbf{84.8} & \textbf{96.5} & \textbf{106.9} \\\hline
			Scenarios & \multicolumn{6}{c|}{Waiting} & \multicolumn{6}{c|}{Walking Dog} & \multicolumn{6}{c|}{Walking Together} & \multicolumn{6}{c}{Average} \\ \hline
			Model & 80ms & 320ms & 560ms & 720ms & 880ms & 1000ms & 80ms & 320ms & 560ms & 720ms & 880ms & 1000ms & 80ms & 320ms & 560ms & 720ms & 880ms & 1000ms & 80ms & 320ms & 560ms & 720ms & 880ms & 1000ms \\ \hline
			Zero-Vel & 21.8 & 72.4 & 104.6 & 117.1 & 125.8 & 130.3 & 37.0 & 99.6 & 126.7 & 140.9 & 154.9 & 160.8 & 26.5 & 85.9 & 116.5 & 121.8 & 123.1 & 122.7 & 23.8 & 76.0 & 107.4 & 121.6 & 131.6 & 136.6 \\
			Res. Sup. & 23.8 & 75.8 & 105.4 & 117.3 & 128.1 & 135.4 & 36.4 & 99.1 & 128.7 & 141.1 & 155.3 & 164.5 & 20.4 & 59.4 & 80.2 & 87.3 & 92.8 & 98.2 & 25.0 & 77.0 & 106.3 & 119.4 & 130.0 & 136.6 \\
			convSeq2Seq & 14.6 & 58.1 & 87.3 & 100.3 & 110.7 & 117.7 & 27.7 & 90.7 & 122.4 & 133.8 & 151.1 & 162.4 & 15.3 & 53.1 & 72.0 & 77.7 & 82.9 & 87.4 & 16.6 & 61.4 & 90.7 & 104.7 & 116.7 & 124.2 \\
			LTD & 10.4 & 47.9 & 77.2 & 90.6 & 101.1 & 108.3 & 22.8 & 77.2 & 107.8 & 120.3 & 136.3 & 146.4 & 10.3 & 39.4 & 56.0 & 60.3 & 63.1 & 65.7 & 12.2 & 50.7 & 79.6 & 93.6 & 105.2 & 112.4 \\
			HRI & 8.7 & \underline{43.4} & 74.5 & 89.0 & 100.3 & 108.2 & 20.1 & \underline{73.3} & 108.2 & 120.6 & 135.9 & 146.9 & 8.9 & \underline{35.1} & \underline{52.7} & \underline{57.8} & \underline{62.0} & \underline{64.9} & 10.4 & \underline{47.1} & 77.3 & 91.8 & 104.1 & 112.1 \\
			PGBIG & \underline{8.4} & \textbf{42.4} & \textbf{71.0} & \textbf{84.6} & \underline{95.6} & \underline{103.2} & \underline{19.9} & \textbf{72.8} & \underline{105.5} & \underline{119.4} & \underline{135.5} & \underline{146.1} & \underline{8.8} & \textbf{35.4} & 54.4 & 61.0 & 64.8 & 67.4 & \underline{10.3} & \textbf{46.6} & \underline{76.3} & \underline{90.9} & \underline{102.6} & \underline{110.0} \\
			Ours & \textbf{7.9} & 46.2 & \underline{74.1} & \underline{84.8} & \textbf{93.4} & \textbf{101.4} & \textbf{18.9} & 74.7 & \textbf{101.9} & \textbf{111.5} & \textbf{126.3} & \textbf{139.6} & \textbf{8.5} & 36.0 & \textbf{48.5} & \textbf{49.9} & \textbf{53.3} & \textbf{57.9} & \textbf{9.9} & 48.8 & \textbf{73.7} & \textbf{84.0} & \textbf{94.3} & \textbf{103.3} \\
			\bottomrule
		\end{tabular}
	}
	\caption{Comparison with deterministic models on Human3.6M~\cite{h36m} Setting-B in FDE (mm) at different prediction horizons in different actions. The best results are highlighted in bold, and the second-best ones are marked with underscores.}
	\label{tab:h36_det_details}
\end{table*}

\end{enumerate}

\end{document}